\documentclass{article}
\usepackage{ijcai22}

\usepackage{times}

\usepackage{soul}
\usepackage{url}
\usepackage[hidelinks]{hyperref}
\usepackage[utf8]{inputenc}
\usepackage[small]{caption}
\usepackage{graphicx}
\usepackage{amsmath}
\usepackage{booktabs}
\usepackage{multicol}
\usepackage{multirow}
\usepackage{makecell}
\usepackage{threeparttable}
\usepackage{natbib}

\title{A Survey of Knowledge-Intensive NLP with Pre-Trained Language Models}

\author{
Da Yin$^1$\footnote{The work was mainly done during internship at Microsoft.}\and
Li Dong$^2$\and
Hao Cheng$^2$\and
Xiaodong Liu$^2$\and
Kai-Wei Chang$^1$\and
Furu Wei$^2$\and
Jianfeng Gao$^2$\\
\affiliations
$^1$University of California, Los Angeles\\
$^2$Microsoft Research\\
\emails
\{da.yin, kwchang\}@cs.ucla.edu,
\{lidong1, chehao, xiaodl, fuwei, jfgao\}@microsoft.com
}

\begin{document}

\maketitle

\begin{abstract}
With the increasing of model capacity brought by pre-trained language models, there emerges boosting needs for more knowledgeable natural language processing (NLP) models with advanced functionalities including providing and making flexible use of encyclopedic and commonsense knowledge. The mere pre-trained language models, however, lack the capacity of handling such knowledge-intensive NLP tasks alone. To address this challenge, large numbers of pre-trained language models augmented with external knowledge sources are proposed and in rapid development. In this paper, we aim to summarize the current progress of pre-trained language model-based knowledge-enhanced models (PLMKEs) by dissecting their three vital elements: knowledge sources, knowledge-intensive NLP tasks, and knowledge fusion methods. Finally, we present the challenges of PLMKEs based on the discussion regarding the three elements and attempt to provide NLP practitioners with potential directions for further research.
\end{abstract}

\section{Introduction}
%Pre-trained language models~\citep{devlin2019bert,radford2019language,DBLP:journals/corr/abs-1907-11692,raffel2020exploring} have achieved enormous advances for natural language processing (NLP). Pre-trained on massive amount of text, the models are able to provide powerful text representations and gain successes on a wide range of NLP tasks~\cite{wang2018glue,wang2019superglue}. However, the machines merely based on pre-trained language models may not satisfy users' need in real-world applications including information seeking and understanding and responding to the situations in our daily routines~\cite{guu2020realm}. A main reason is that the pre-trained language models themselves lack the help of additional encyclopedic and commonsense knowledge that is relevant to the query or context.

Pre-trained language models~\citep{devlin2019bert,radford2019language} have achieved enormous advances for natural language processing (NLP).
Using various types of self-supervised language modeling objectives over massive text corpora, the resulting models provide powerful text representations for supervised fine-tuning on downstream tasks, leading to great successes on a wide range of NLP tasks~\citep{wang2018glue,wang2019superglue}.
In particular, recent study shows that certain knowledge (linguistic or factual knowledge \citep{Manning2020linguistic,petroni-etal-2019-language,roberts-etal-2020-much,knowledgeneurons}) is implicitly stored in their parameters which partially explains the better generalization abilities of NLP models based on pre-trained language models.
However, the machines merely based on the \textit{implicit knowledge} from pre-trained language models pose challenges for dynamically updating learned knowledge (e.g., correcting biases or enriching knowledge sources), which may not satisfy users' need in real-world applications, such as serving our daily information seeking routines~\citep{guu2020realm}.
In other words, one main drawback of existing pre-trained language models is the lack of ability to leverage \textit{explicit} encyclopedic and commonsense knowledge.

The growing real-world needs form the motivation to investigate \textbf{P}re-Trained \textbf{L}anguage \textbf{M}odel-based \textbf{K}nowledge-\textbf{E}nhanced Models (\textbf{PLMKE}s). 
Here, we establish the framework of PLMKE for NLP, involving a knowledge interface responsible for fetching relevant \emph{external knowledge} with text inputs and 
%a reasoning module performs a joint inference over both natural language inputs and retrieved explicit knowledge data.
a \emph{knowledge fusion module} integrating knowledge with the representations learned from pre-trained language models.
%, and potential \emph{knowledge-intensive applications} used to customize models.
%Incorporating pre-trained language models as foundations, PLMKEs, strengthened by external knowledge sources, are aimed for tackling NLP tasks requiring human knowledge. Specifically, given a text input of a particular NLP task, PLMKEs attempt to leverage the relevant knowledge with the input from knowledge sources to make predictions. 
Recently, PLMKEs have been widely used in various knowledge-intensive tasks including open-domain question answering~\citep{guu2020realm,DBLP:conf/nips/LewisPPPKGKLYR020,cheng-etal-2021-unitedqa}, fact verification~\citep{zhou-etal-2019-gear,liu-etal-2020-fine}, entity linking~\citep{jiang-etal-2021-exploring-listwise} and commonsense reasoning~\citep{lin-etal-2019-kagnet,yasunaga-etal-2021-qa}.

Given that different tasks require various kinds of knowledge, most recent PLMKEs customize the source for knowledge interface and the design of corresponding knowledge fusion method. Thus, we structure our survey so that different knowledge-intensive scenarios can be understood through our lens.
Specifically, there are three vital items related to PLMKEs:

\paragraph{Knowledge Sources:} Knowledge sources (Wikipedia, \citep{bollacker2008freebase,vrandevcic2012wikidata,DBLP:conf/aaai/SpeerCH17,sap2019atomic,Zhang2020TransOMCSFL}, etc.) provide external knowledge to PLMKEs and lay the foundations of PLMKEs along with pre-trained language models. The contents of knowledge sources are also decisive to the tasks PLMKEs can solve.

\paragraph{Knowledge-Intensive NLP Tasks:} Knowledge-intensive NLP tasks~(\citep{nist2004,thorne-etal-2018-fever,kwiatkowski-etal-2019-natural,petroni2021kilt}, etc.) are testbeds to evaluate the performance of PLMKEs on checking whether the selected knowledge sources are appropriate and manifest the effectiveness of knowledge fusion methods.

\paragraph{Knowledge Fusion Methods:} Knowledge fusion methods~(\citep{zhang2019ernie,lin-etal-2019-kagnet,guu2020realm,DBLP:journals/corr/abs-2107-02137}, etc.) involve the implementations of exploiting knowledge sources to empower pre-trained language models in PLMKEs and achieve better performance on knowledge-intensive NLP tasks.

In this paper, we aim to provide researchers a general view on PLMKEs. The survey centers on the aforementioned three elements, \textbf{knowledge sources}, \textbf{knowledge-intensive NLP tasks}, and \textbf{knowledge fusion methods}. We will discuss the following questions in the rest of sections.

\begin{itemize}
    \item \textbf{Section~\ref{know_sour} (Knowledge Sources)}: What are the types of commonly used knowledge sources? What is the format of knowledge stored in the knowledge sources? What are the characteristics of the knowledge sources?
    
    \item \textbf{Section~\ref{know_task} (Knowledge-Intensive NLP Tasks)}: What are the common knowledge-intensive NLP tasks PLMKEs are applied on? What knowledge is useful to solve the tasks? 
    %What are the characteristics of these tasks compare with other well-known NLP tasks?
    
    \item \textbf{Section~\ref{know_meth} (Knowledge Fusion Methods)}: How can we categorize the numerous knowledge fusion methods? What do the different categories of fusion methods typically do when fusing knowledge? What are fusion methods adopted in common PLMKEs and why?
    
    \item \textbf{Section~\ref{challenges} (Challenges and Future Directions)}: What are the potential challenges of PLMKEs? What are the future directions to solve these challenges?
\end{itemize}

We notice that there are two contemporaneous surveys~\citep{wei2021knowledge,yang2021survey} about knowledge-enhanced NLP models. %\cite{wei2021knowledge,yang2021survey} 
However, they both focus on the models that incorporate knowledge during pre-training stage, which are a subset of the scope that our survey studies. Besides, we adopt a novel taxonomy to dissect and categorize PLMKEs and propose new perspectives for the future directions upon the taxonomy.

\section{Knowledge Sources}
\label{know_sour}
Knowledge sources provide pre-trained language models with needed knowledge and empower them with higher capability to handle knowledge-intensive NLP tasks. We list the common knowledge sources leveraged in PLMKEs in Table~\ref{knowledge}, and further categorize them into two groups: encyclopedic knowledge and commonsense knowledge.

\subsection{Encyclopedic Knowledge}
\textbf{Encyclopedic knowledge} contains attributes (e.g., \textsf{age}, \textsf{duration}) about entities (e.g., \textsf{person}, \textsf{event}) and the relations (e.g., \textsf{educated at}, \textsf{followed by}) between entities. Wikipedia is one of the prevalent knowledge sources providing massive amount of encyclopedic knowledge including biography of a person and background of an event. Extracted from unstructured text corpora, factual knowledge bases aim at uncovering graph structures of entities and converting them into structured database. Typically, structured encyclopedic knowledge is represented by triplets containing entity names and their relationships (e.g., \textless\textsf{Tom Hanks, occupation, actor}\textgreater). Factual knowledge bases (e.g., Wikidata) are also widely used in PLMKEs~\citep{peters2019knowledge,agarwal-etal-2021-knowledge}.

\begin{table}[!ht]
\centering
\scriptsize
\begin{tabular}{lcc}
\toprule
\textbf{Knowledge Types} & \textbf{Knowledge Sources} & \textbf{Knowledge Domains} \\
\midrule
& \makecell[c]{Wikipedia/Wikidata \\ \cite{vrandevcic2012wikidata}}       &   \\
\cmidrule(lr{1em}){2-2}
& \makecell[c]{DBPedia \\ \cite{auer2007dbpedia}}       & \multirow{3}{*}[1em]{open domain} \\
\cmidrule(lr{1em}){2-2}
\multirow{5}{*}[1.5em]{\textbf{Encyclopedic Knowledge}} & \makecell[c]{Freebase \\ \cite{bollacker2008freebase}}       &  \\
\cmidrule(lr{1em}){2-3}
& \makecell[c]{UMLS \\ \cite{bodenreider2004unified}}       & biomedicine  \\
\cmidrule(lr{1em}){2-3}
& \begin{tabular}[c]{@{}c@{}}AMiner \\ \cite{Tang2008ArnetMinerEA} \end{tabular}       & science \\
\cmidrule(lr{1em}){1-3}
& \makecell[c]{ConceptNet \\ \cite{DBLP:conf/aaai/SpeerCH17}}       &  \\
\cmidrule(lr{1em}){2-2}
& \makecell[c]{TransOMCS \\ \cite{Zhang2020TransOMCSFL}}       & \multirow{3}{*}[1em]{open domain} \\
\cmidrule(lr{1em}){2-2}
\multirow{6}{*}[1.5em]{\textbf{Commonsense Knowledge}} & \makecell[c]{CSKG \\ \cite{DBLP:conf/esws/IlievskiSZ21}}       &  \\
\cmidrule(lr{0.5em}){2-3}
& \makecell[c]{ATOMIC \\ \cite{sap2019atomic}}       & \multirow{2}{*}[-1em]{human interaction} \\
\cmidrule(lr{1em}){2-2}
& \makecell[c]{ATOMIC$^{20}_{20}$ \\ \cite{Hwang2021COMETATOMIC2O}}       &   \\
\cmidrule(lr{1em}){2-3}
& \makecell[c]{ASER \\ \cite{DBLP:conf/www/ZhangLPSL20}}       & eventuality \\
\bottomrule
\end{tabular}
\caption{Common knowledge sources used in PLMKEs.}
\label{knowledge}
\end{table}

\subsection{Commonsense Knowledge}
\textbf{Commonsense knowledge} includes the basic facts about situations in human's daily life. It involves everyday events and their effects (e.g., \textsf{mop up the floor if we split food over it}), facts about beliefs and desires (e.g., \textsf{study hard to win scholarship}), and properties of objects (e.g., \textsf{goat has four legs}). Thus, different from encyclopedic knowledge, commonsense knowledge is usually shared by most people and implicitly assumed in communications. 

Similar to the storage method of factual knowledge bases, commonsense knowledge sources also use triplets to represent knowledge. These sources depict commonsense including subtype relationship between objects (e.g., \textless\textsf{apple, IsA, fruit}\textgreater in ConceptNet), cause and effect of an event (e.g., \textless\textsf{personX adopts a pet, Effects, play with the pet}\textgreater), and intent of human behaviour (e.g., \textless\textsf{personX adopts a pet, Causes, to have a companion}\textgreater). We can observe that the main difference from factual knowledge bases is that the triplets contain everyday objects and their elements are typically described with a short sentence. Recent PLMKEs~\citep{lin-etal-2019-kagnet,mitra2019additional} mostly utilize the sources including ConceptNet and ATOMIC as external knowledge to enhance models' commonsense reasoning capacity.

\subsection{Characteristics of Current Knowledge Sources}
We discuss two typical characteristics of current knowledge sources: \textbf{large-scale} and \textbf{diverse}. Regarding to the scale of knowledge sources, all the encyclopedic knowledge sources in Table~\ref{knowledge} contain millions of concepts and at least hundred million of facts induced by them. The largest commonsense sources in Table~\ref{knowledge} is ASER, which contains 64 million facts. We observe that the size of commonsense knowledge sources is much smaller than the encyclopedic ones. However, compared with prior commonsense sources such as Cyc~\citep{lenat1995cyc}, the current sources are produced in more precise and scalable way: the annotation process is partially automatic and accessible to non-experts. The current trend of commonsense collection methods manifests the potential to scale up the knowledge sources.

The domains that common knowledge sources cover are diverse. Encyclopedic knowledge sources such as Wikipedia, DBPedia and Freebase are collected from open domain, suggesting that they are constructed by heterogeneous knowledge not limited to specific domains. Meanwhile, knowledge sources involving specific domains such as biomedicine and science (e.g., UMLS and AMiner) are established to boost the development of domain-specific applications. Since commonsense involves various aspects including human interaction and object properties in everyday life, there exist both open-domain commonsense knowledge sources (e.g., ConceptNet, TransOMCS) that cover multiple domains of commonsense, and domain-specific commonsense sources (e.g., ATOMIC, ASER) that focus on particular types. The diversity of knowledge sources is beneficial for laying the foundations of broader future applications of PLMKEs.

\section{Knowledge-Intensive NLP Tasks}
\label{know_task}
Knowledge-intensive NLP tasks are served as testbeds to evaluate the capability of PLMKEs to solve problems that require external knowledge. In this section, we provide an overview of knowledge-intensive NLP tasks, and summarize their corresponding features. 

\subsection{Overview of Knowledge-Intensive NLP Tasks}
Knowledge-intensive NLP tasks can be divided into two groups based on the types of the required knowledge: encyclopedic and commonsense knowledge-intensive tasks. 

Table~\ref{ency_tasks} lists several typical datasets of three representative \textbf{encyclopedic knowledge-intensive tasks}: open-domain question answering (QA), fact verification, and entity linking. The open-domain QA task aims to answer information seeking questions (e.g., \textsf{``When was Barack Obama born?''}) that require encyclopedic knowledge over various domains. Fact verification (e.g., judge if the claim \textsf{``Barack Obama born was born on August 4 1961.''} is true) is designed to verify whether a given text claim is factually correct which also demands the model's ability to reason over large amount of factual knowledge. Lastly, the purpose of entity linking system is to link text mentions of entities to their corresponding unique identifier in a target database, e.g., linking to Wikipedia pages. The datasets~\citep{thorne-etal-2018-fever,kwiatkowski-etal-2019-natural,DBLP:conf/cikm/GuoB14} for these tasks heavily relies on encyclopedic knowledge sources including Wikipedia and DBPedia.

\textbf{Commonsense knowledge-intensive tasks} focus on testing whether models can accurately understand and respond to daily scenarios. For example, in a task about social commonsense, given a context like \textsf{``Someone spilled the food all over the floor''}, models are required to select the most proper responses like \textsf{``He/she needs to mop up''} instead of unreasonable ones like \textsf{``Run around in the mess''}. The types of commonsense-intensive tasks are diverse because of the diversity of commonsense knowledge. As shown in Table~\ref{common_tasks}, tasks about social commonsense involve human interactions and thoughts; when it comes to physical commonsense, the questions in the datasets inquire physical properties and ways to manipulate objects; temporal commonsense reasoning datasets usually contain questions about event order, duration and frequency.

\begin{table}[!ht]
\centering
\scriptsize
\scalebox{0.92}{
\begin{tabular}{ccc}
\toprule
\textbf{Tasks} & \textbf{Datasets} & \textbf{Data Sources} \\
\midrule
& \begin{tabular}[c]{@{}c@{}}\textsc{Natural Questions}\\ \cite{kwiatkowski-etal-2019-natural}\end{tabular}       & Wikipedia  \\
\cmidrule(lr{1em}){2-3}
\multirow{2}{*}[2em]{\makecell[c]{\textbf{Open-domain QA}}} & \begin{tabular}[c]{@{}c@{}}\textsc{HotpotQA} \\ \cite{yang-etal-2018-hotpotqa}\end{tabular}       & Wikipedia \\
\cmidrule(lr{1em}){1-3}
& \begin{tabular}[c]{@{}c@{}}\textsc{FEVER}\\ \cite{thorne-etal-2018-fever}\end{tabular}       & Wikipedia  \\
\cmidrule(lr{1em}){2-3}
\multirow{2}{*}[2em]{\makecell[c]{\textbf{Fact Verification}}} & \begin{tabular}[c]{@{}c@{}}\textsc{BoolQ} \\ \cite{clark-etal-2019-boolq}\end{tabular}       & Wikipedia \\
\cmidrule(lr{1em}){1-3}
& \begin{tabular}[c]{@{}c@{}}\textsc{ACE2004} \\ \cite{nist2004}\end{tabular}       & \makecell[c]{news}  \\
\cmidrule(lr{1em}){2-3}
\multirow{4}{*}[0.3em]{\makecell[c]{\textbf{Entity Linking}}} & \begin{tabular}[c]{@{}c@{}}\textsc{AIDA CoNLL-YAGO} \\ \cite{hoffart-etal-2011-robust}\end{tabular}       & \makecell[c]{DBPedia \& YAGO \\ \cite{suchanek2007yago}} \\
\cmidrule(lr{1em}){2-3}
& \begin{tabular}[c]{@{}c@{}}\textsc{WnWi}\\ \cite{DBLP:conf/cikm/GuoB14}\end{tabular}       & Wikipedia  \\
\cmidrule(lr{1em}){2-3}
& \begin{tabular}[c]{@{}c@{}}\textsc{WnCw} \\ \cite{DBLP:conf/cikm/GuoB14}\end{tabular}       & Clueweb~\footnotemark[1]  \\
\bottomrule
\end{tabular}
}
\caption{Detailed information about representative encyclopedic knowledge-intensive tasks and datasets.}
\label{ency_tasks}
\end{table}
\footnotetext[1]{\url{https://lemurproject.org/clueweb12/}}

\begin{table}[!ht]
\centering
\scriptsize
\scalebox{0.92}{
\begin{tabular}{ccc}
\toprule
\textbf{Commonsense Types} & \textbf{Tasks/Datasets} & \textbf{Data Sources}  \\
\midrule
& \begin{tabular}[c]{@{}c@{}}\textsc{CommonsenseQA} \\ \cite{talmor-etal-2019-commonsenseqa}\end{tabular}       & ConceptNet  \\
\cmidrule(lr{1.2em}){2-3}
\multirow{4}{*}[0em]{\makecell[c]{\textbf{General Commonsense}}} & \begin{tabular}[c]{@{}c@{}}\textsc{WSC} \\ \cite{levesque2012winograd}\end{tabular}       & human thoughts \\
\cmidrule(lr{1em}){2-3}
& \begin{tabular}[c]{@{}c@{}}\textsc{$\alpha$NLI}\\ \cite{bhagavatula2020abductive}\end{tabular}       & \makecell[c]{ROCStories \\ \cite{mostafazadeh-etal-2016-corpus}}  \\
\cmidrule(lr{1em}){2-3}
& \begin{tabular}[c]{@{}c@{}}\textsc{CommonGen} \\ \cite{lin-etal-2020-commongen}\end{tabular}       & \makecell[c]{image captions \& \\ ConceptNet} \\
\cmidrule(lr{1em}){1-3}
\makecell[c]{\textbf{Social Commonsense}} & \begin{tabular}[c]{@{}c@{}}\textsc{SocialIQA} \\ \cite{sap-etal-2019-social}\end{tabular}       & \makecell[c]{ATOMIC \\ \cite{sap2019atomic}} \\
\cmidrule(lr{1em}){1-3}
\multirow{2}{*}[-0.7em]{\makecell[c]{\textbf{Physical Commonsense}}} & \begin{tabular}[c]{@{}c@{}}\textsc{PIQA} \\ \cite{Bisk2020}\end{tabular}       & \makecell[c]{image captions \& \\ ConceptNet} \\
\cmidrule(lr{1em}){2-3}
& \begin{tabular}[c]{@{}c@{}}\textsc{HellaSWAG} \\ \cite{zellers-etal-2019-hellaswag}\end{tabular}       & video captions  \\
\cmidrule(lr{1em}){1-3}
%\begin{tabular}[c]{@{}c@{}}\textsc{VCR} \\ \cite{zellers2019vcr}\end{tabular}       & Movies & visual commonsense \\
%\cmidrule(lr{1em}){1-3}
\multirow{2}{*}[-0.7em]{\makecell[c]{\textbf{Temporal Commonsense}}} & \begin{tabular}[c]{@{}c@{}}\textsc{McTaco} \\ \cite{zhou-etal-2019-going}\end{tabular}       & \makecell[c]{MultiRC \\ \cite{khashabi-etal-2018-looking}}  \\
\cmidrule(lr{1em}){2-3}
& \begin{tabular}[c]{@{}c@{}}\textsc{TACRIE} \\ \cite{zhou-etal-2021-temporal}\end{tabular}       & \makecell[c]{ROCStories \\ \cite{mostafazadeh-etal-2016-corpus}}  \\
\bottomrule
\end{tabular}
}
\caption{Detailed information about types of commonsense and representative commonsense knowledge-intensive tasks.}
\label{common_tasks}
\end{table}

\subsection{Characteristics of Knowledge-Intensive Tasks}

The most salient characteristic is that external encyclopedic or commonsense knowledge is necessary to perform well on these tasks. Not only for models, it is hard for humans to answer questions about Barack Obama's birth date without any reference knowledge. Another characteristic is that although these tasks can be tackled with external knowledge, the required knowledge may not be directly given along with input. In other words, the default task input is simply a question or context of a certain scenario, without any additional information. It motivates researchers to consider adding a module in charge of grounding to external knowledge sources in the design of PLMKEs.

\section{Knowledge Fusion Methods}
\label{know_meth}
Tackling knowledge-intensive NLP tasks highly relies on the assistance of appropriate knowledge sources. How to fuse such important knowledge into the strong pre-trained language models and make them more knowledgeable poses challenges to researchers. 

The training process of pre-trained language models usually involve two stages: pre-training and fine-tuning~\citep{devlin2019bert}. Pre-training is a self-supervised learning process to learn representations through language modeling on large unlabeled text corpus. Fine-tuning is the process to adapt pre-trained models with task-specific supervision on downstream tasks. In PLMKEs, knowledge can be integrated in either stage but the fusion methods for the two stages are quite different. In this section, we categorize the mainstream knowledge fusion methods by the stage where knowledge is fused in PLMKEs.
%, and sum up the findings based on empirical results of the methods.

\begin{figure}[htbp]
\centering
\includegraphics[width=0.39\textwidth, trim=0 70 75 0, clip]{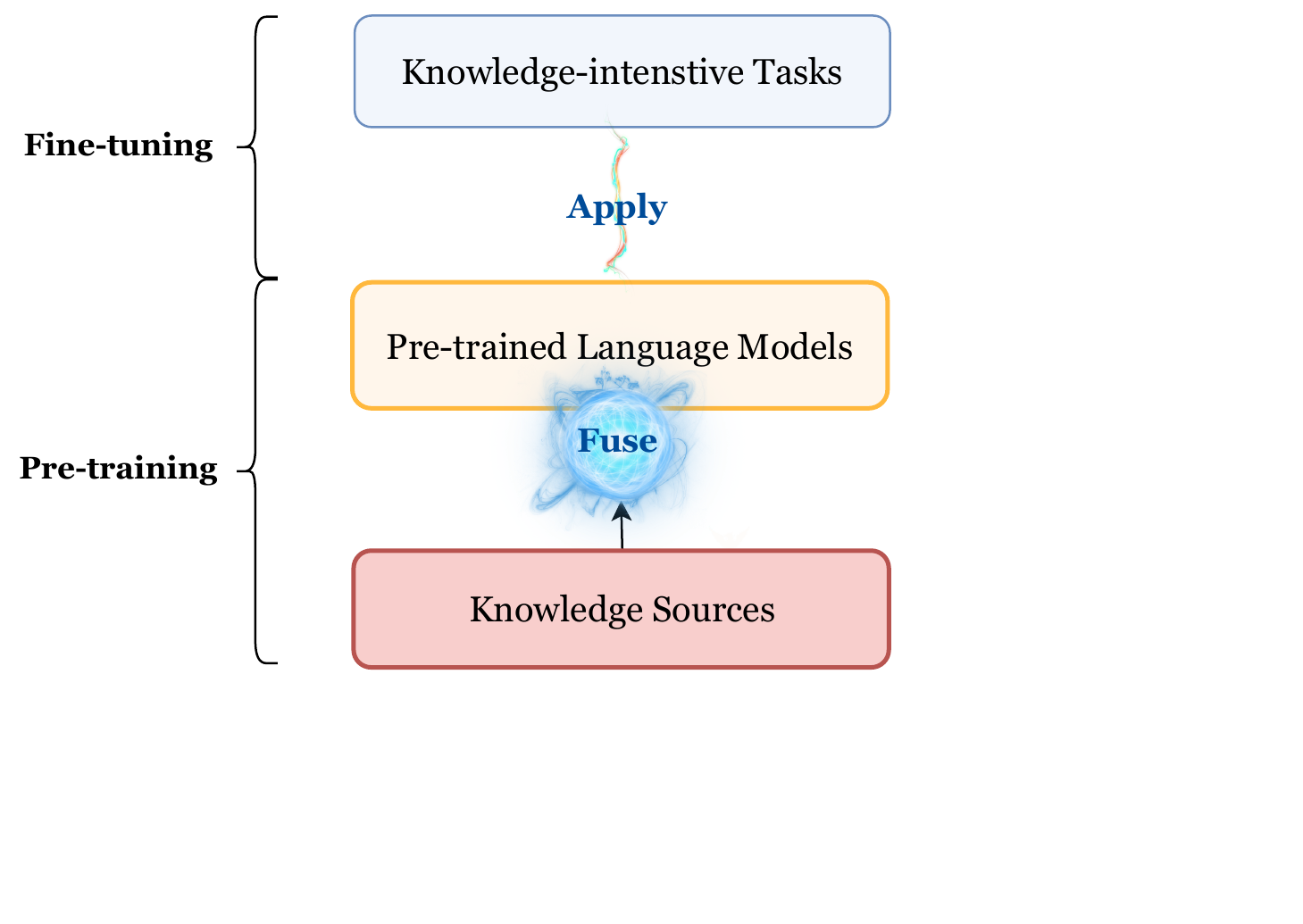}
\caption{Pre-fusion methods.}
\label{pre-fusion}
\end{figure}

\subsection{Pre-Fusion Methods}
Pre-fusion methods fuse external knowledge in the \textbf{pre-training stage}~\citep{zhang2019ernie,DBLP:journals/corr/abs-2107-02137}. Before knowledge fusion, knowledge sources are first processed into the format similar to unstructured raw text corpus. Then, the processed text corpus are used for further pre-training with a sharing set of learning objectives used by original language models. Thus, pre-fusion methods enable knowledge fusion without much architectural change. For knowledge sources such as knowledge graphs, however, the knowledge is usually structured and not aligned with the unstructured input format of language models. The simplest approach to tackling this challenge is concatenating the entities and relation~\citep{zhang2019ernie} or generating fluent synthetic sentences by conditional text generation models~\citep{agarwal-etal-2021-knowledge}.

\begin{table*}[]
\centering
\small
\scalebox{0.92}{
\begin{tabular}{lcccc}
\toprule
\textbf{Tasks}          & \textbf{Datasets}                                         & \textbf{Models}                   & \textbf{Fusion Types}                   & \textbf{Fused Knowledge}        \\
\midrule
\multirow{5}{*}[-3.7em]{\textbf{Open-domain QA}} & \makecell[c]{\textsc{Natural Questions}\\ \cite{kwiatkowski-etal-2019-natural}} & \makecell[c]{UnitedQA \\ \cite{cheng-etal-2021-unitedqa}}                 & Post-fusion                    &  Wikipedia                      \\ \cmidrule(lr{1em}){2-5}
                                & \makecell[c]{\textsc{WebQuestion}\\ \cite{berant-etal-2013-semantic}}                                               & \makecell[c]{EMDR$^2$ \\ \cite{DBLP:journals/corr/abs-2106-05346}} & Post-fusion & Wikipedia \\ \cmidrule(lr{1em}){2-5}
                                & \makecell[c]{\textsc{TriviaQA}\\ \cite{joshi-etal-2017-triviaqa}}                                               & \makecell[c]{EMDR$^2$ \\ \cite{DBLP:journals/corr/abs-2106-05346}} & Post-fusion & Wikipedia \\
                                \cmidrule(lr{1em}){2-5}
                                & \multirow{2}{*}[-1em]{\makecell[c]{Other Representative Models}}              
                                &  \makecell[c]{REALM \\ \cite{guu2020realm}} & Hybrid-fusion & Wikipedia \\
                                \cmidrule(lr{1em}){3-5}
                                &                                                                   & \makecell[c]{RAG \\ \cite{DBLP:conf/nips/LewisPPPKGKLYR020}} & Hybrid-fusion & Wikipedia \\
                                \midrule
\multirow{4}{*}[-2.7em]{\textbf{Fact Verification}} & \makecell[c]{\textsc{FEVER}\\ \cite{thorne-etal-2018-fever}} & \makecell[c]{\citet{jiang-etal-2021-exploring-listwise}}               & Post-fusion                   & Wikipedia                       \\ \cmidrule(lr{1em}){2-5}
                                & \makecell[c]{\textsc{BoolQ}\\ \cite{berant-etal-2013-semantic}}                                               & \makecell[c]{ERNIE 3.0 \\ \cite{DBLP:journals/corr/abs-2107-02137}} & Pre-fusion & \makecell[c]{Wikipedia, Bookcorpus \\ \cite{zhu2015aligning}, etc.} \\ \cmidrule(lr{1em}){2-5}
                             %   &                                                                   & \makecell[c]{FiD-Ex \\ \cite{DBLP:journals/corr/abs-2012-15482}} & Post-fusion & Wikipedia \\
                             %   \cmidrule(lr{1em}){2-5}
                                & \multirow{2}{*}[-0.7em]{\makecell[c]{Other Representative Models}}
                                & \makecell[c]{GEAR \\ \cite{zhou-etal-2019-gear}} & Post-fusion & Wikipedia \\
                                \cmidrule(lr{1em}){3-5}
                                &                                                                   & \makecell[c]{KGAT \\ \cite{liu-etal-2020-fine}} & Post-fusion & Wikipedia \\
                             %   \cmidrule(lr{1em}){3-5}
                             %   &                                                                   & \makecell[c]{Transformer-XH \\ \cite{DBLP:conf/iclr/ZhaoXRSBT20}} & Post-fusion & Wikipedia \\
                             %   \cmidrule(lr{1em}){3-5}
                             %   &                                                                   & \makecell[c]{DREAM \\ \cite{zhong-etal-2020-reasoning}} & Post-fusion & Wikipedia \\
                                \midrule
\multirow{2}{*}[-0.7em]{\textbf{Entity Linking}} & \makecell[c]{\textsc{AIDA CoNLL-YAGO} \\ \cite{hoffart-etal-2011-robust}} & \citet{mulang2020evaluating}               & Post-fusion                   & Wikipedia                        \\ \cmidrule(lr{1em}){2-5}
& \makecell[c]{Other Representative Models}
                                & \makecell[c]{CHOLAN \\ \cite{ravi2021cholan}} & Post-fusion & Wikipedia \\
                     %           & \multirow{2}{*}[-1em]{\makecell[c]{\textsc{BoolQ}\\ \cite{berant-etal-2013-semantic}}}                                               & \makecell[c]{ERNIE 3.0 \\ \cite{DBLP:journals/corr/abs-2107-02137}} & Pre-fusion & Wikipedia, Bookcorpus, etc. \\ \cmidrule(lr{1em}){3-5}
                     %           &                                                                   & \makecell[c]{FiD-Ex \\ \cite{DBLP:journals/corr/abs-2012-15482}} & Post-fusion & Wikipedia \\
%                                \midrule
%\multirow{4}{*}[-1em]{\makecell[c]{\textbf{Knowledge Triplet} \\ \textbf{Slot Filling}}} & \textsc{Google-RE}\tnote{1} & \makecell[c]{KELM \\ \cite{agarwal-etal-2021-knowledge}}               & Pre-fusion                   & Wikipedia \& WebNLG                       \\ \cmidrule(lr{1em}){2-5}
%                                & \multirow{2}{*}[-1em]{\makecell[c]{\textsc{T-Rex}\\ \cite{elsahar-etal-2018-rex}}}                                        
%                                & \makecell[c]{KELM \\ \cite{agarwal-etal-2021-knowledge}} & Pre-fusion & Wikipedia \& WebNLG \\ \cmidrule(lr{1em}){3-5}
%                                &                                                                   & \makecell[c]{KGI$_0$ \\ \cite{DBLP:journals/corr/abs-2104-08610}} & Post-fusion & Wikipedia \\
\bottomrule
\end{tabular}
}
\caption{State-of-the-art PLMKEs on encyclopedic knowledge-intensive tasks and other representative models.}
\label{ency_plmkes}
\end{table*}

\begin{figure}[htbp]
\centering
\includegraphics[width=0.46\textwidth, trim=0 165 130 3, clip]{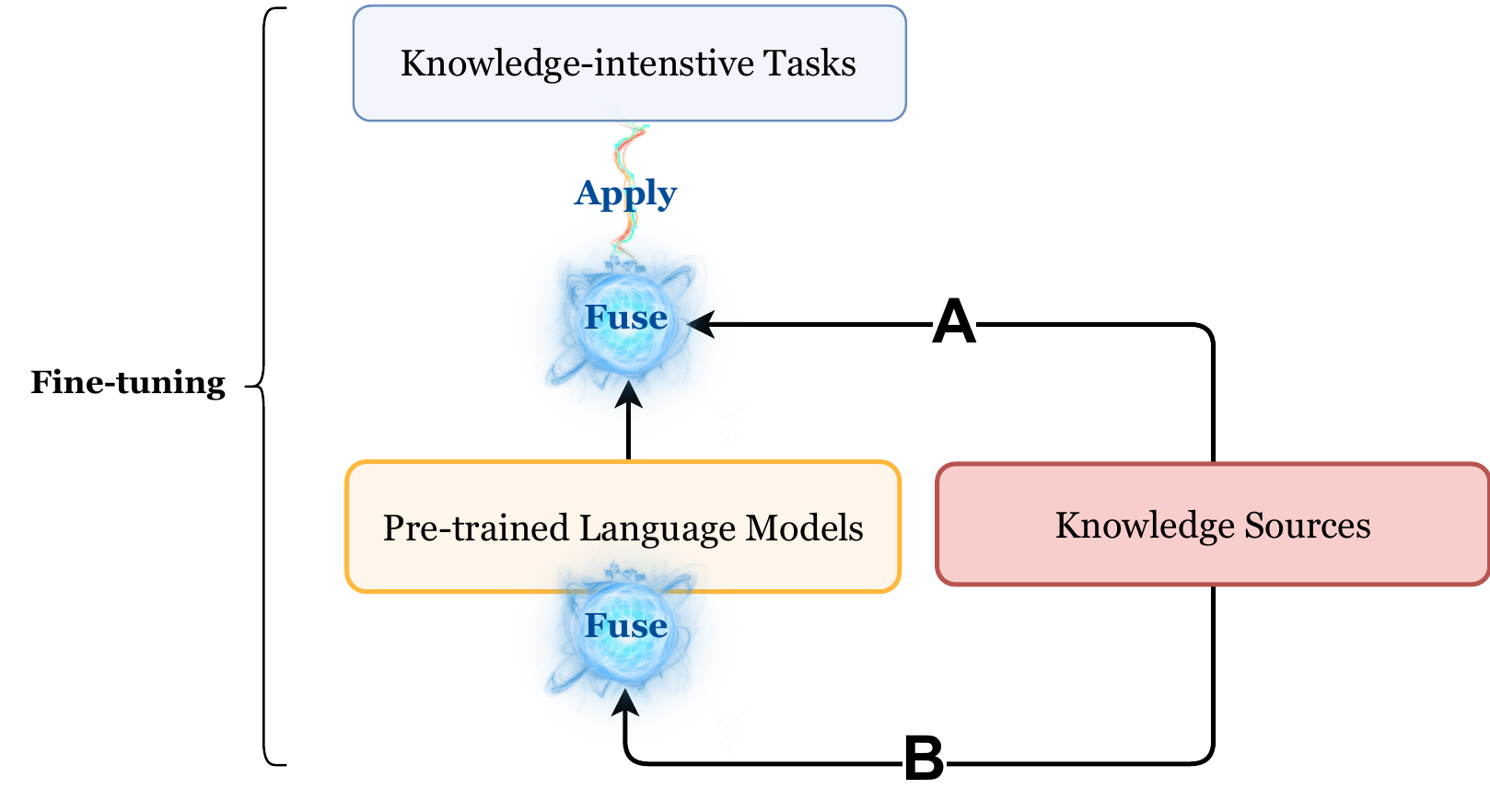}
\caption{Post-fusion methods. Path A indicates structure-aware post-fusion methods that encode structured knowledge into supplemental embeddings and integrate with the text embeddings produced by pre-trained language models. Path B indicates text-based post-fusion methods that retrieve relevant texts and subgraphs and convert them into texts to be fed into pre-trained language models. Commonly, each PLMKE chooses either path A or B to fuse knowledge.}
\label{post-fusion}
\end{figure}

\subsection{Post-Fusion Methods}
Instead of incorporating knowledge in the pre-training stage, post-fusion methods seek to fuse the knowledge in the \textbf{fine-tuning stage}~\citep{zhou-etal-2019-gear,lin-etal-2019-kagnet,liu-etal-2020-fine,cheng-etal-2021-unitedqa}. Given an input text from a particular knowledge-intensive task, post-fusion methods first retrieve the relevant knowledge to the input, and then perform a joint reasoning on top of the augmented input. 

To capture relevant knowledge, the post-fusion methods leverage text retriever on unstructured knowledge sources to extract the useful textual spans; for structured knowledge sources, previous works attempt to match the entities appearing in input text to the relevant entity-centric subgraphs. After capturing the knowledge, they select one of the two following approaches to implementing knowledge fusion. The captured knowledge can be transformed into knowledge embeddings by encoders such as graph neural networks, and used as supplemental features to the text embeddings provided by pre-training language models for the following reasoning modules (structure-aware post-fusion: path A in Figure~\ref{post-fusion})~\citep{lin-etal-2019-kagnet,yasunaga-etal-2021-qa}. It can also be directly concatenated with the text input and fed into the pre-training language models altogether (text-based post-fusion: path B in Figure~\ref{post-fusion})~\citep{karpukhin2020dense,cheng-etal-2021-unitedqa}.

\subsection{Hybrid-Fusion Methods}
Hybrid-fusion methods are a combination of pre-fusion and post-fusion methods: knowledge is fused in both \textbf{pre-training} and \textbf{fine-tuning stages}. Although there exists salient difference between pre-fusion and post-fusion methods, the hybrid-fusion methods enable us to unify both: the retriever frequently leveraged in post-fusion methods can be jointly trained in the pre-training stage. That is, during the pre-training, language models are also learning to leverage additional retrieved knowledge for modeling the language context. The pre-trained model augmented by the jointly learned retriever can thus utilize the knowledge from the retriever more effectively during the fine-tuning stage. The retrieval-augmented pre-training~\citep{peters2019knowledge,guu2020realm,DBLP:conf/nips/LewisPPPKGKLYR020} is commonly adopted in hybrid-fusion methods and it manifests the effectiveness on several knowledge-intensive tasks discussed in Section~\ref{know_task}.

\begin{figure}[htbp]
\centering
\includegraphics[width=0.4\textwidth, trim=0 135 200 3, clip]{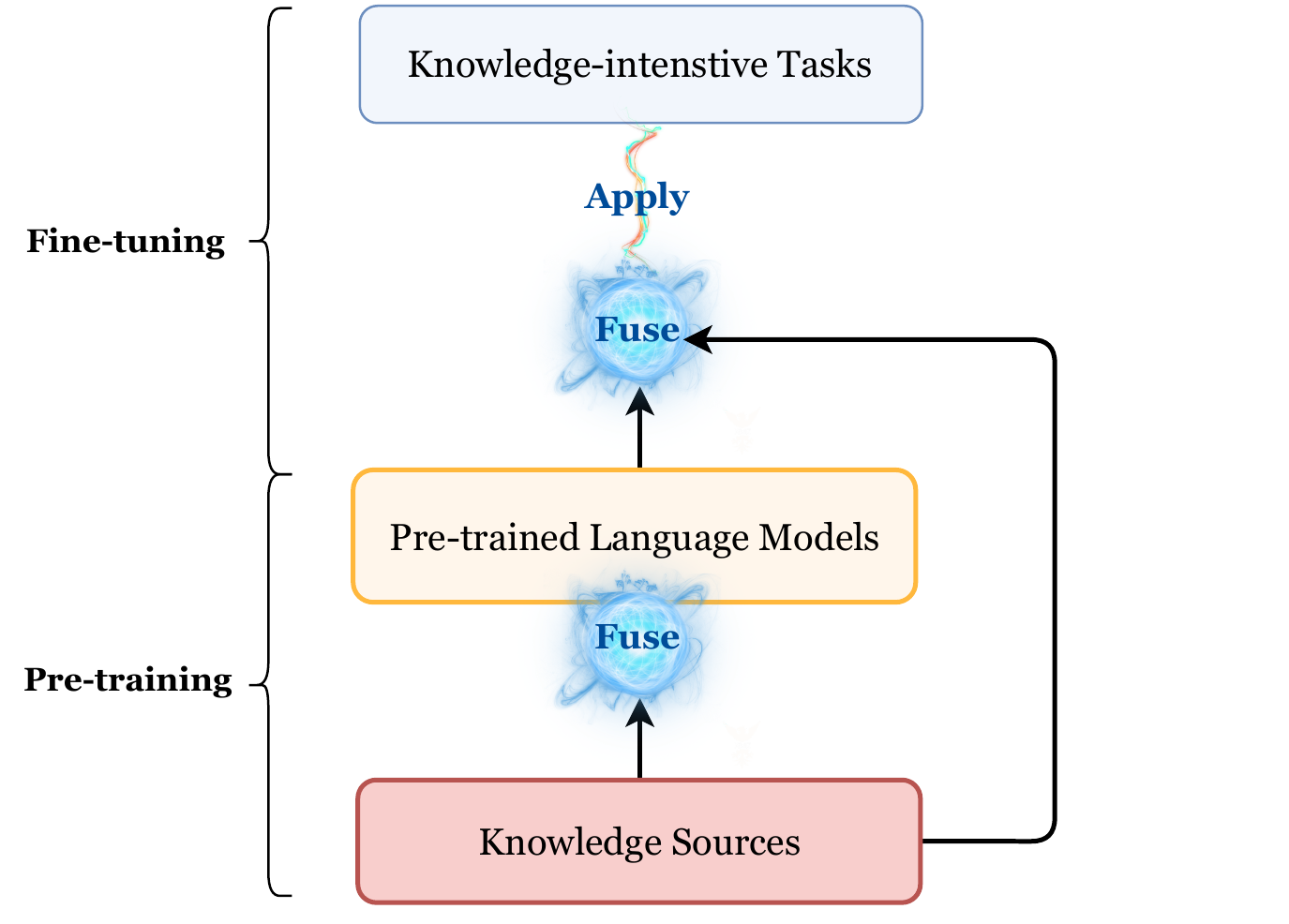}
\caption{Hybrid-fusion methods.}
\label{hybrid-fusion}
\end{figure}

\begin{table*}[]
\centering
\small
\scalebox{0.92}{
\begin{tabular}{lcccc}
\toprule
\textbf{Commonsense Types}          & \textbf{Tasks/Datasets}                                         & \textbf{Models}                   & \textbf{Fusion Types}                   & \textbf{Fused Knowledge}        \\
\midrule
\multirow{5}{*}[-3.7em]{\textbf{General Commonsense}} & \makecell[c]{\textsc{CommonsenseQA}\\ \cite{talmor-etal-2019-commonsenseqa}} & \makecell[c]{GreaseLM \\ \cite{zhang2022greaselm}}                 & Post-fusion                    &  ConceptNet                      \\ \cmidrule(lr{1em}){2-5}
                                & \makecell[c]{\textsc{WSC}\\ \cite{levesque2012winograd}}                                               & \makecell[c]{ERNIE 3.0 \\ \cite{DBLP:journals/corr/abs-2107-02137}} & Pre-fusion & \makecell[c]{Wikipedia, Bookcorpus \\ \cite{zhu2015aligning}, etc.} \\ \cmidrule(lr{1em}){2-5}
                                & \makecell[c]{\textsc{$\alpha$NLI} \\ \cite{bhagavatula2020abductive}}                                               & \makecell[c]{UNIMO \\ \cite{li-etal-2021-unimo}} & Pre-fusion & Wikipedia, Bookcorpus, images \\
                                \cmidrule(lr{1em}){2-5}
                                & \multirow{2}{*}[-0.7em]{\makecell[c]{Other Representative Models}}              
                                &  \makecell[c]{KagNet \\ \cite{lin-etal-2019-kagnet}} & Post-fusion & ConceptNet \\
                                \cmidrule(lr{1em}){3-5}
                                &                                                                   & \makecell[c]{QA-GNN \\ \cite{yasunaga-etal-2021-qa}} & Post-fusion & ConceptNet \\
                                \midrule
\multirow{3}{*}[-1.5em]{\textbf{Social Commonsense}} & \makecell[c]{\textsc{SocialIQA}\\ \cite{sap-etal-2019-social}} & \makecell[c]{UNICORN \\ \cite{lourie2021unicorn}}               & Pre-fusion                   & various commonsense benchmarks                       \\ \cmidrule(lr{1em}){2-5}
                             %   &                                                                   & \makecell[c]{FiD-Ex \\ \cite{DBLP:journals/corr/abs-2012-15482}} & Post-fusion & Wikipedia \\
                             %   \cmidrule(lr{1em}){2-5}
                                & \multirow{2}{*}[-0.7em]{\makecell[c]{Other Representative Models}}
                                & \makecell[c]{UnifiedQA-11B \\ \cite{khashabi-etal-2020-unifiedqa}} & Pre-fusion & various QA benchmarks \\
                                \cmidrule(lr{1em}){3-5}
                                &                                                                   & \makecell[c]{McQueen \\ \cite{mitra2019additional}} & Post-fusion & ATOMIC \\
                             %   \cmidrule(lr{1em}){3-5}
                             %   &                                                                   & \makecell[c]{Transformer-XH \\ \cite{DBLP:conf/iclr/ZhaoXRSBT20}} & Post-fusion & Wikipedia \\
                             %   \cmidrule(lr{1em}){3-5}
                             %   &                                                                   & \makecell[c]{DREAM \\ \cite{zhong-etal-2020-reasoning}} & Post-fusion & Wikipedia \\
                                \midrule
\multirow{2}{*}[-0.7em]{\textbf{Physical Commonsense}} & \makecell[c]{\textsc{PIQA} \\ \cite{Bisk2020}} & \makecell[c]{UNICORN \\ \cite{lourie2021unicorn}}               & Pre-fusion                   & various commonsense benchmarks                       \\ \cmidrule(lr{1em}){2-5}
%   &                                                                   & \makecell[c]{FiD-Ex \\ \cite{DBLP:journals/corr/abs-2012-15482}} & Post-fusion & Wikipedia \\
%   \cmidrule(lr{1em}){2-5}
& Other Representative Models & \makecell[c]{UnifiedQA-11B \\ \cite{khashabi-etal-2020-unifiedqa}} & Pre-fusion & various QA benchmarks \\
%\\ \cmidrule(lr{1em}){2-5}
                     %           & \multirow{2}{*}[-1em]{\makecell[c]{\textsc{BoolQ}\\ \cite{berant-etal-2013-semantic}}}                                               & \makecell[c]{ERNIE 3.0 \\ \cite{DBLP:journals/corr/abs-2107-02137}} & Pre-fusion & Wikipedia, Bookcorpus, etc. \\ \cmidrule(lr{1em}){3-5}
                     %           &                                                                   & \makecell[c]{FiD-Ex \\ \cite{DBLP:journals/corr/abs-2012-15482}} & Post-fusion & Wikipedia \\
%                                \midrule
%\multirow{4}{*}[-1em]{\makecell[c]{\textbf{Knowledge Triplet} \\ \textbf{Slot Filling}}} & \textsc{Google-RE}\tnote{1} & \makecell[c]{KELM \\ \cite{agarwal-etal-2021-knowledge}}               & Pre-fusion                   & Wikipedia \& WebNLG                       \\ \cmidrule(lr{1em}){2-5}
%                                & \multirow{2}{*}[-1em]{\makecell[c]{\textsc{T-Rex}\\ \cite{elsahar-etal-2018-rex}}}                                        
%                                & \makecell[c]{KELM \\ \cite{agarwal-etal-2021-knowledge}} & Pre-fusion & Wikipedia \& WebNLG \\ \cmidrule(lr{1em}){3-5}
%                                &                                                                   & \makecell[c]{KGI$_0$ \\ \cite{DBLP:journals/corr/abs-2104-08610}} & Post-fusion & Wikipedia \\
\bottomrule
\end{tabular}
}
\caption{State-of-the-art PLMKEs on commonsense knowledge-intensive tasks and other representative models.}
\label{commonsense_plmkes}
\end{table*}

\subsection{Representative Models for Specific Tasks}

Here, we first categorize representative models for various knowledge-intensive NLP tasks based on their corresponding knowledge fusion methods discussed in previous section.
Table~\ref{ency_plmkes} and \ref{commonsense_plmkes} list state-of-the-art (SOTA) PLMKEs on encyclopedic and commonsense knowledge-intensive tasks and several representative models, respectively. For encyclopedic knowledge-intensive tasks, it is shown that except \textsc{BoolQ}, the other state-of-the-art models all adopt post-fusion methods. On the contrary, for commonsense knowledge-intensive tasks, except \textsc{CommonsenseQA}, pre-fusion methods are broadly leveraged in the SOTA models. 

We then analyze why pre-fusion methods are not prevalent and effective on encyclopedic knowledge-intensive tasks. In pre-fusion methods, the knowledge required in these tasks is implicitly stored in pre-trained parameters. But it is hard to determine what the knowledge is finally stored, and it also increases the difficulty in eliciting and leveraging the knowledge~\citep{guu2020realm,wang2021can}. Instead, post-fusion methods are able to infer upon explicit and concrete textual knowledge. But the advantage of post-fusion to leverage explicit and concrete knowledge may become a shortcoming for commonsense knowledge-intensive tasks. As mentioned in Section~\ref{know_sour}, commonsense is usually implicit inside texts and the coverage of commonsense knowledge sources is much smaller than that of encyclopedic knowledge sources. Even if we compose large-scale commonsense knowledge sources with the help of knowledge acquisition methods, we are still likely to miss a large body of commonsense knowledge used in our daily life. Therefore, it is possible that post-fusion methods may fail at retrieving the relevant knowledge inside the knowledge sources and thus cannot bring extra benefits to the pre-trained language models.

\section{Challenges and Future Directions}
\label{challenges}

% In this section, we present the challenges that PLMKEs face and propose potential directions that support further development of PLMKEs.

\subsection{Unified PLMKEs Across Tasks and Domains}

Recent developments of PLMKEs have led to task-specific modeling advances. As shown in Table~\ref{ency_plmkes}, the models frequently used on encyclopedic knowledge-intensive tasks adopt post-fusion and hybrid-fusion methods, while the two fusion methods are not usually exploited on commonsense knowledge-intensive tasks. Furthermore, we observe that the state-of-the-art models in different knowledge-intensive NLP tasks are unique, making the progress on each task seemingly incompatible. Beyond the tasks listed in Table~\ref{ency_tasks} and~\ref{common_tasks}, knowledge-intensive NLP tasks are extended to various domains involving biomedical and legal~\citep{liu2021everything} knowledge. Recently, researchers also attach more importance to the diversity of knowledge existing in different times~\citep{dhingra2021time} and regions~\citep{zhang2021situatedqa,yin2021broaden,liu2021visually}. The diversity across tasks and domains is naturally fostering the need for unified PLMKEs, instead of promoting the trend of devising unique models on individual tasks. 

\subsection{Reliabilty of Knowledge Sources}
Since knowledge sources are the basis of PLMKEs, we are concerned with the reliability of knowledge sources. Currently, many large-scale knowledge sources are constructed by automatic knowledge acquisition algorithms. Whereas, there exists a trade-off between the scale and precision: it is likely to introduce false and biased information into the knowledge sources~\citep{sun2021men}. We anticipate bias amplification in case the PLMKEs are constructed on biased knowledge sources. We call upon researchers to be aware of the reliability of knowledge sources via proposing more precise knowledge acquisition algorithms and careful inspection over the knowledge sources they intend to use.

\subsection{Reasoning Module Design}
Reasoning is an important step for solving knowledge-intensive NLP tasks. It is essential especially for commonsense knowledge-intensive tasks, since the relevant commonsense knowledge is usually implicit and should be used in multiple turns of reasoning. When we humans encounter the situation like \textsf{``Someone spilled the food all over the floor''}, we are first aware of the fact that the floor is not clean, and others' shoes would get dirty if they stepped on the spilled food. Based on the situation, the intent to mop up the floor produces. Though several PLMKEs achieve great performance on such commonsense-related scenarios, it is unclear about whether the models perform human-like reasoning implicitly or simply capture the spurious correlation and thus not robust upon more complex situations. Existing models containing reasoning modules mainly perform reasoning on entity~\citep{lin-etal-2019-kagnet,yasunaga-etal-2021-qa} or syntax structures~\citep{yin2020sentibert,bai2021syntax}, which cannot cover the complex situations like the aforementioned \textsf{``spill food''} example. To achieve the human-like capability of recognizing the everyday situations, multi-hop reasoning module is needed for designing trustworthy PLMKEs that can simulate human thoughts.

\section{Conclusions}
We comprehensively survey existing works about knowledge-intensive NLP with pre-trained language models and summarize the current progress in terms of the three critical components in PLMKEs: knowledge sources, knowledge-intensive NLP tasks, and knowledge fusion methods. Based on the discussion about the three components, we further pose several challenges that would be influential in the practical usage and propose the related future directions in response to the challenges. We hope that this paper could provide NLP practitioners with a clear picture on the topic and boost the development of the current knowledge-intensive NLP technologies.

\clearpage

\bibliography{openqa,fact_ver,ent_link,triple,commonsense,know_source,fact_methods,others}
\bibliographystyle{named}

\end{document}